\documentclass{article}

\usepackage{microtype}
\usepackage{graphicx}
\usepackage[dvipsnames,table]{xcolor}
\usepackage{subfigure}
\usepackage{booktabs} 
\usepackage{amsmath,amssymb,amsfonts}
\usepackage{tabularx}

\usepackage{hyperref}

\newcommand{\highlight}[1]{\textbf{#1}}

\usepackage[accepted]{icml2021}

\icmltitlerunning{Evaluating the progress of Deep Reinforcement Learning in the real world}

\begin{document}

\twocolumn[
\icmltitle{Evaluating the progress of Deep Reinforcement Learning in the real world: aligning domain-agnostic and domain-specific research
}




\begin{icmlauthorlist}
\icmlauthor{Juan Jose Garau-Luis}{esl}
\icmlauthor{Edward Crawley}{esl}
\icmlauthor{Bruce Cameron}{esl}
\end{icmlauthorlist}

\icmlaffiliation{esl}{Engineering Systems Laboratory, Massachusetts Institute of Technology, Cambridge, MA, USA}
\icmlcorrespondingauthor{Juan Jose Garau-Luis}{garau@mit.edu}

\icmlkeywords{Deep Reinforcement Learning, Real World, Domain-agnostic, Domain-specific}

\vskip 0.3in
]



\printAffiliationsAndNotice{}  

\begin{abstract}
Deep Reinforcement Learning (DRL) is considered a potential framework to improve many real-world autonomous systems; it has attracted the attention of multiple and diverse fields. Nevertheless, the successful deployment in the real world is a test most of DRL models still need to pass. In this work we focus on this issue by reviewing and evaluating the research efforts from both domain-agnostic and domain-specific communities. On one hand, we offer a comprehensive summary of DRL challenges and summarize the different proposals to mitigate them; this helps identifying five gaps of domain-agnostic research. On the other hand, from the domain-specific perspective, we discuss different success stories and argue why other models might fail to be deployed. Finally, we take up on ways to move forward accounting for both perspectives.
\end{abstract}

\section{Introduction}
\label{sec:introduction}

In the recent years, multiple research fields and industries have become interested in Deep Reinforcement Learning (DRL) frameworks as a way to enhance decision-making processes in the real world and design better autonomous systems. The range of domains is large \cite{Li2017a,Naeem2020}, including---but not limited to---robotics \cite{Polydoros2017}, communications \cite{Luong2019}, drug discovery \cite{Elton2019}, fluid mechanics \cite{GARNIER2021104973}, autonomous vehicles \cite{Talpaert2019}, and recommender systems \cite{Afsar2021}. 

Different motivations trigger the interest of domain-specific research in DRL:
\begin{itemize}
    \item In some industries systems are scaling and presenting new degrees of freedom, which makes them harder to operate, especially in time-sensitive settings. In those cases DRL becomes a new approach to fast decision-making. An example is the work within the satellite communications community, in a moment when constellations are getting larger and more flexible \cite{Deng2020,Ferreira2019a}.
    \item Some systems require control policies that leverage raw signals such as image, sound, or brain activity. DRL and its representation capabilities are thus studied to achieve better performance. Some robotics \cite{Polydoros2017,Ibarz2021} and healthcare applications \cite{Esteva2019,Yu2019} fall in this category.
    \item When supervision is costly or not possible, DRL offers a way to learn policies by encoding system goals into the reward function and leveraging exploration during training. This is the case of NP-hard combinatorial optimization problems \cite{Drori2020} or drug design studies \cite{Olivecrona2017,Popova2018}, where DRL is able to provide approximate solutions and good candidate molecules, respectively.
    \item DRL is a framework that can account for long-term dependencies in decision-making, which is especially interesting for recommender platforms and other interaction-based systems \cite{Afsar2021,Zhang2019}. 
\end{itemize}

Despite these motivations and the extensive research efforts to prove the usefulness of DRL in real-world contexts, the successful deployment in real environments is a test many of the proposed models in the literature still need to pass. This is mostly due to the added complexities of the real world compared to current experimental DRL settings.

From a domain-specific perspective, concrete tasks, problems, and environments in the real world are hard to fully characterize and training in real environments is not always possible or preferred. In addition, the nature of these tasks or problems---regardless of the domain---entails dealing with certain challenges that make learning more difficult: non-stationarity, high-dimensionality, sparse reward, etc.

Reinforcement Learning (RL) researchers have identified a large number of those challenges \cite{dulac2019challenges,Dulac-Arnold2021,Ibarz2021,Zhu2020} and many solutions have been proposed to mitigate each of them. While the results are positive, testing is mostly limited to simulated environments. A key question remains often unanswered: how do the proposed models work in a real-world setting? Without that feedback it is not clear how much we have progressed in the path towards DRL-based autonomy.

To try to shed some light on this issue, in this paper we attempt to summarize and evaluate the progress of real-world-oriented DRL research from the perspective of both domain-agnostic and domain-specific research. We start by reviewing the domain-agnostic challenges of real-world DRL and compiling the solutions proposed in the literature. Based on this review, we identify five gaps that we deem necessary to address moving forward: a bias towards robotics use cases, not enough research on combined challenges, lack of real-world follow-through, little understanding of design tradeoffs, and operation being ignored.

In addition to the focus of the RL community on the problem, we argue domain-specific operators play an important role in adopting the technology. Therefore, we next analyze the same problem---the lack of real-world deployments---from the domain-specific viewpoint and try to align both efforts. We first highlight some examples of success stories, then discuss why other deployments might fail, and finally address ways to move forward.

\section{Challenges of real-world DRL}
\label{sec:challenges}

\begin{figure}[t]
\centering
\includegraphics[width=.99\linewidth]{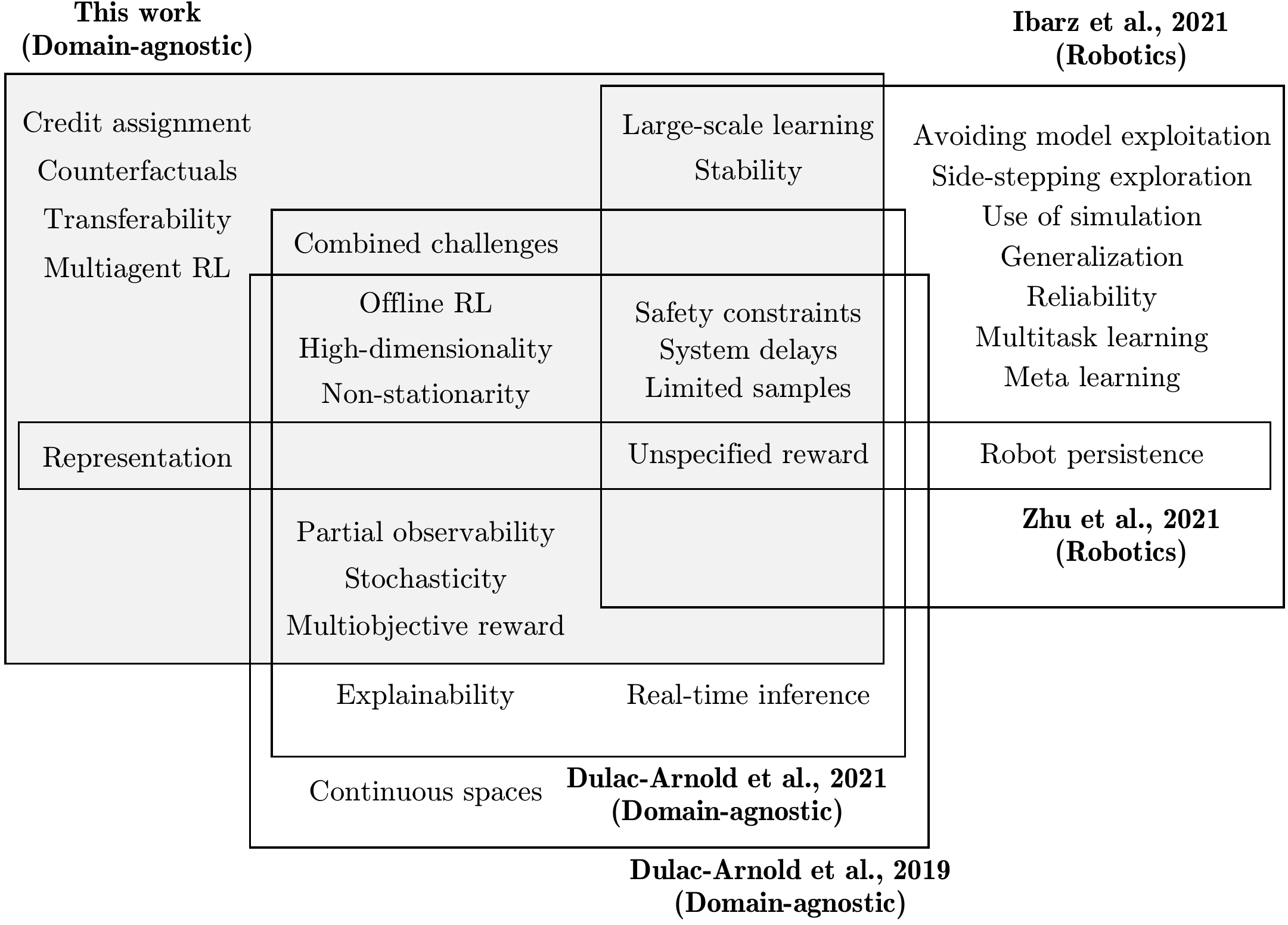}
\caption{List of challenges of real-world DRL considered in different studies. Some of the challenges might interact or overlap in specific settings.}
\label{fig:challenges}
\end{figure}

Different studies have tried to summarize the challenges of real-world DRL, these are pictured in Figure \ref{fig:challenges}. The work in \cite{dulac2019challenges,Dulac-Arnold2021} offers a comprehensive review of nine different domain-agnostic challenges and their impact on trained agents. Within the context of robotics, \cite{Ibarz2021} identifies twelve issues based on real case studies and discusses possible mitigation strategies for each. Also focusing on robotics, \cite{Zhu2020} highlights three real-world challenges and proposes a model taking those into account on a set of dexterous robotic manipulation tasks. 

While the set of challenges is diverse, the mitigation strategies to address them are not unique to one challenge but sometimes the same method or mechanism is proposed in different contexts. To summarize the data, in this work we assume a domain-agnostic perspective and extend the challenges identified in \cite{Dulac-Arnold2021}. In the following section, we address each challenge independently and list the mitigation approaches found in the literature. 

The challenges we consider are: offline DRL, learning from limited samples, large-scale learning, high-dimensionality, safe DRL, partial observability, non-stationarity, unspecified reward, multiobjective reward, system delays, representation, transferability, long trajectories and credit assignment, stochasticity, multiagent DRL, counterfactual reasoning, stability, and the combination of many of these challenges.

Note these challenges do not necessarily need to be uncorrelated, in some contexts specific challenges might be consequences of other challenges being present (e.g., non-stationarity during learning might be a cause of a partially observable environment). There are two other challenges some authors have highlighted that are not included in our analysis as isolated challenges: continuous spaces and real-time inference. In our view these challenges are hard to be found being addressed in isolation and therefore are excluded from our list.

Finally, designing, implementing, and operating a DRL model in the real world might also entail other considerations that go beyond difficult learning setups and are directly connected with the societal implications of the technology. Here we refer to interpretability (identified as one of the nine challenges in \cite{Dulac-Arnold2021}), reproducibility (discussed in detail in \cite{Henderson2018}), reliability, fairness, and privacy, as other challenges that are out of the scope of this work but remain important on a societal level.

\section{Challenge mitigation strategies}
\label{sec:strategies}

In this section we aim to summarize the approaches different researchers have proposed to address the DRL challenges introduced in the previous section. Table \ref{tab:summary} offers a summary of the high-level approaches and the contexts in which they are applied. This summary is based on our view of the challenges and the reviewed literature; it is open to different interpretations.

\renewcommand{\arraystretch}{1.4}
\begin{table*}[t]
\caption{Summary of the different approaches that have been considered by the RL community to address the specific challenges of real-world DRL. Each approach has been considered for different challenges independently.}
\label{tab:summary}
\vskip 0.1in
\begin{center}
\begin{small}
\begin{tabular}{p{0.15\linewidth-2\tabcolsep-1.3333\arrayrulewidth}p{0.4\linewidth-2\tabcolsep-1.3333\arrayrulewidth}p{0.45\linewidth-2\tabcolsep-1.3333\arrayrulewidth}}
\hline
\multicolumn{1}{c}{\textbf{Approach}} & \multicolumn{1}{c}{\textbf{Description}}                                       & \multicolumn{1}{c}{\textbf{Examples}}                                                                                      \\ \hline
\rowcolor[HTML]{EFEFEF}
Meta learning  & An outer loop learner changes meta parameters to better adapt to the challenge & Meta learning for offline DRL, meta learning for multiobjective reward, Meta RL for transferability, replacing action maximization by neural network search in high-dimensional spaces \\
Mathematical guarantee & Derive equations and theorems that support the challenge fulfillment & Lyapunov functions and primal-dual methods for safe DRL, assume uncertainty matrices to address non-stationarity \\
\rowcolor[HTML]{EFEFEF}
Neural network architectures & Rely on Deep Learning advances to increase robustness against the challenge & RNNs for partial observability, attention mechanisms for credit assignment, network ensembles for offline DRL \\
RL theory & Adapt theoretical RL frameworks to DRL settings and the use of neural networks & Off-policy algorithms, POMDPs, Delay-Aware MDPs, Constrained MDPs, Maximum-entropy RL \\
\rowcolor[HTML]{EFEFEF}
Embeddings and latent spaces & Address challenge problems by relying on robust intermediate embeddings & High- to low-dimensional embeddings, latent variables for non-stationarity, multimodal and contrastive learning-based representations, unsupervised learning \\
Reward estimation or modification & Try to overcome the challenge by directly modifying the reward structure and/or function & Reward shaping in long trajectories, reward shaping for safe DRL, reward redistribution, distributional RL against stochasticity \\
\rowcolor[HTML]{EFEFEF}
Deriving a model & Instead of learning a policy, learn models of the environment and use them to plan & Model-based RL, imitation learning, inverse RL \\
Pruning and masking & The learning process involves deciding, among different learning signals, how important each of them is and eliminating the unnecessary ones & Batch-constrained methods for offline DRL, distributed training in large-scale settings, action elimination in high-dimensional spaces, divide-and-conquer methods in stochastic environments \\
\rowcolor[HTML]{EFEFEF}
Use auxiliary tasks & Provide the agent with auxiliary tasks that, combined, increase robustness against the challenge & Multitask learning and online learning for data efficiency, self-supervised learning for unspecified reward, hierarchical RL in long trajectories \\
Data augmentation & Rely on different data augmentation and data wrangling techniques & Randomization to transfer better, data augmentation to address non-stationarity and stochasticity \\
\rowcolor[HTML]{EFEFEF}
Heuristics & Use human-crafted rules or processes to address the challenge & Scalarization of multiple objectives, hyperparameter tuning \\
Population-based methods & Have multiple agents with slightly different parameters/objectives and search for the best ones & Multiagent populations in partially observable environments, multiobjective populations \\
\rowcolor[HTML]{EFEFEF}
Multiagent specific & Solutions specific to the multiagent challenge that can not be mapped to other challenges & Independent Q-learning, decentraliced actors and centralized critic, hybrid mechanisms \\
\hline
\end{tabular}
\end{small}
\end{center}
\vskip -0.1in
\end{table*}

\textbf{Offline DRL}
\,\,\, Some systems might require to learn from offline logs instead of directly interacting with the environment, as that might be costly or not possible. An extensive review on the subject is presented in \cite{levine2020offline}. To address this issue, different \highlight{off-policy algorithms} such as DDPG \cite{lillicrap2015continuous} or D4PG \cite{Barth-Maron2018} can be used in some cases. Other authors propose \highlight{batch-constrained RL} approaches \cite{fujimoto2019off,Kumar2019,Siegel2020,Wu2019}, where the learned policy is constrained based on the state-action distribution from the dataset and the extrapolation error is accounted for. Then, the work in \cite{agarwal2020optimistic} considers an \highlight{ensemble} or convex combination of Q-functions to leverage data in a replay buffer. Finally, \highlight{model-based RL} \cite{sutton2018reinforcement} constitutes another research area in the context of offline DRL \cite{Yu2020a,Kidambi2020}.

\textbf{Learning from limited samples}
\,\,\, Sometimes an agent must learn from a small number of samples, either because acquiring experience is slow or costly, or rapid adaptation to a new context is needed. While the representations chosen or learned can impact the learning speed \cite{Srinivas2020}, multiple methods have been proposed to directly address data efficiency. One alternative is to \highlight{learn a model} of the world and use that model to plan \cite{chua2018deep,buckman2018sample}. In the context of learning specific tasks, if \highlight{expert demonstrations} \cite{duan2017one} or behavioral priors \cite{Singh2020} are available, the agent can bootstrap from those to increase data efficiency. If the goal of the agent is \highlight{multitask learning}, the tasks can be learned concurrently taking into account multiple gradient inputs \cite{Yu2020}, or if the tasks are to be learned sequentially, \highlight{meta learning} algorithms, especially few-shot methods, offer a way to learn new tasks faster \cite{Finn2017,Li2017,sung2018learning,lee2019meta}. Finally, in \highlight{online learning} contexts, where new tasks need to be learned fast and on-the-fly, different approaches have been proposed to promote forward and backward transfer \cite{Schwarz2018,mallya2018packnet,Chaudhry2018,Nagabandi2018} and avoid catastrophic forgetting \cite{kirkpatrick2017overcoming}.

\textbf{Large-scale learning}\,\,\, In specific settings an agent should be able to capitalize on massive amounts of data fast, either because experience comes at a high frequency or multiple independent agents can collect experience simultaneously. For the latter case, when environments can be parallelized (e.g., self-driving cars, recommender systems), \highlight{distributed training with importance and priority mechanisms} has been proposed in different works \cite{Adamski2018,Horgan2018,Espeholt2018}. 

\textbf{High-dimensionality}
\,\,\, Some agents might need to operate in high-dimensional or combinatorial state and action spaces (e.g., natural language, molecular space). Here one approach is to operate with lower dimensional \highlight{embeddings} of the spaces \cite{Dulac-Arnold2015,Robine2020}. Other authors propose \highlight{action elimination} mechanisms to determine which actions not to take first \cite{zahavy2018learn}. In the context, of Q-learning \cite{sutton2018reinforcement} over large action spaces, \cite{van2020q} proposes \highlight{replacing the maximization} operation for a neural network. Finally, the use of \highlight{canonical spaces} can help reduce the state space by encapsulating redundant spaces together \cite{wu2017emergent}. 

\textbf{Safe DRL}
\,\,\, While exploration plays an important part in the success of RL, agents acting in real-world environments should account for safety constraints and be able to evaluate risks. One common approach to that end is to encode constraints as part of the reward function \cite{garcia2015comprehensive}, but that might not be always desirable \cite{achiam2017constrained}. Some studies propose adding learnable safety layer on top of the policy in order to \highlight{prune or correct unsafe actions} \cite{dalal2018safe}. The work in \cite{Tessler2018} explores \highlight{reward shaping} and proposes a method that substracts constraint-violation penalties to the reward. Then, \highlight{Lyapunov functions} have been proposed to certify the stability and safety of different RL-based controllers \cite{Chow2018,berkenkamp2017safe}. Constraint satisfaction can also be guaranteed by means of \highlight{primal-dual methods}, as shown in \cite{qin2021density}. Finally, an agent can also learn to \highlight{trade rewards and costs} by specifying constraints as costs with state-dependent and learnable Lagrangian coefficients \cite{Bohez2019}. 

\textbf{Partial observability}
\,\,\, Many environments in the real world are partially observable. In the context of DRL, some authors initially proposed incorporating \highlight{past observations} to the state \cite{mnih2015human} or use \highlight{recurrent neural network} architectures \cite{Hausknecht2015}. Inspired by the theory on POMDPs \cite{cassandra1994acting}, works like \cite{igl2018deep} propose training a variational autoencoder to learn latent representations encoding \highlight{belief states}. In the case the agent competes against other non-fully-observable agents, \cite{Jaderberg2019} shows that \highlight{training populations of agents} eventually leads to best agents finding suitable policies for the environment. If the agent must cooperate with the other agents, the use of \highlight{shared experience replay} helps mitigating the effect of partial observability \cite{omidshafiei2017deep}.

\textbf{Non-stationarity}
\,\,\, A robust policy should be effective in non-stationary environments, where the underlying transition dynamics change over time due to various factors such as noise. In these contexts, one alternative is the use of \highlight{latent variables} that encode environment representations robust to noisy cues \cite{Xie2020}. A well-established approach is to \highlight{assume uncertainty in the transition matrices and derive robust algorithms} that consider worst-case scenarios \cite{Mankowitz2019} or pursue soft-robustness \cite{Derman2018}. Bayesian optimization-based methods can be also derived from this latter idea \cite{derman2020bayesian}. Finally, \highlight{data augmentation and randomization} during training can also lead to policies that adapt to real-world environments and generalize better \cite{Peng2018}. 

\textbf{Unspecified reward}
\,\,\, Sometimes agents must learn skills without reward signals, due to unavailable human feedback, complex exploration dynamics, or long horizon tasks. If there is no reward function but expert demonstrations are available, \highlight{Inverse RL} is an approach to learn reward signals \cite{Fu2017}. The work in \cite{Hansen2020} proposes a method to train policies by means of \highlight{self-supervised learning} when deploying in environments without reward information. Another alternative is to learn a goal-conditioned policy via \highlight{unsupervised learning}, maximizing the similarity between visited states and a goal state \cite{Warde-Farley2018}. In the context of multitask learning, in \cite{Eysenbach2018} an agent is shown to learn a diverse set of distinguishable skills by \highlight{maximizing entropy}. These skills can be then used to beetter adapt to new tasks. 

\textbf{Multiobjective reward}
\,\,\, Several tasks in the real world require accounting for multiple objectives and an agent must learn to reason about them. To that end, many works rely on \highlight{scalarization} approaches that combine the different objectives into a weighted reward function. This approach can be hard to tune if there are changes on the individual rewards' scale or their priorities over time. To have a better control over the objectives, \cite{abdolmaleki2020distributional} proposes training \highlight{individual policies for each objective} and then, instead of combining rewards in the reward space, combine policies in the distribution space. Another alternative is to train a different \highlight{policy per preference over objectives} \cite{xu2020prediction,Yang2019}, which leads to dense Pareto-optimal sets of policies that trade the different objectives following the operator's preferences. Finally, \highlight{meta learning} methods have also been proposed to learn new preferences in a few-shot fashion \cite{Chen2019}. 

\textbf{System delays}
\,\,\, DRL experimental setups generally assume negligible delay when acting, observing the new state, or receiving the reward. That might not be the case in the real world. To address this issue, the framework of \highlight{Delay-Aware MDPs} was introduced in \cite{Chen2020} to account for delayed dynamics. A similar idea is proposed in \cite{Derman2021}, where the delayed-Q algorithm leverages a forward dynamics model to predict delayed states. In the context of recommendation systems, the method in \cite{Mann2018} exploits \highlight{intermediate observations/symbols} to mitigate the effect of delays.

\textbf{Representation}\,\,\, In certain environment the challenge lies in encoding all information relevant to the problem or task efficiently, leveraging the sufficient amount of domain knowledge or inductive biases \cite{Hessel2019}. Tradeoffs are present, e.g., learning policies from physical state-based features might be more sample-efficient---although not always possible---than learning from pixels \cite{Tassa2018}. The question ``what makes a good representation for RL?'' is studied in \cite{Singh2020}. A simple approach is to design different representations for the same environment and turn the specific chosen representation into a \highlight{hyperparameter} that can be tuned based on the scenario \cite{Kim2020}. Different environment encodings can be also combined into \highlight{multimodal representations} (e.g., image and sound in video-based environments) \cite{Tsai2018,Tian2019}. Helpful representations can be also learned, for instance by means of \highlight{contrastive learning} frameworks \cite{Wu2018,Srinivas2020}. Then, \cite{Zhang2020a} proposes learning \highlight{invariant representations} by means of lossy autoencoders that capture only task-relevant elements. Finally, representation problems can be also regarded from the perspective of the reward; better reward functions might be devised following \highlight{reward shaping} methods \cite{Faust2019,Chiang2019}.

\textbf{Transferability}\,\,\, Policies should be able to be transferred to different instances of the system and/or environment regardless of their low-level features, without posing a considerable challenge. To that end, \highlight{randomization} strategies can be used to increase robustness against transfer \cite{Lee2019,Tobin2017}. \highlight{Meta RL} methods serve as another way to achieve transferability, by parametrizing specific elements of the DRL framework and using an outer loop learner trained on multiple environments \cite{Oh2020,houthooft2018evolved,Kirsch2019,Alet2020}. Learning common \highlight{invariant latent spaces} could be another approach to consider in some contexts \cite{gupta2017learning}.

\textbf{Long trajectories and credit assignment}\,\,\, When trajectories are long and/or rewards are sparse, learning efficient behaviors can be hard; the agent must discover a long sequence of correct actions. \highlight{Hierarchical RL} poses a possible solution, by considering a hierarchy of auxiliary tasks with known reward structure in order for the agent to reason at different levels of temporal resolution \cite{Riedmiller2018,nachum2018data,vezhnevets2017feudal}. Other works propose \highlight{attention mechanisms} to ease credit assignment over long timescales \cite{Wayne2018,Hung2019}. Then, the method presented in \cite{arjona2019rudder} tackles the problem by \highlight{redistributing reward}, i.e., creating a return-equivalent MDP that redistributes reward more uniformly. Finally, \highlight{reward shaping} methods are also studied for this type of contexts \cite{Su2015,Chiang2019}.

\textbf{Stochasticity}
\,\,\, In some occasions, real-world environments can be too stochastic, which might lead to high variance gradient estimates that hamper learning. To make sure the agent is trained over a wide distribution of states, using \highlight{data augmentation} strategies and \highlight{randomization} are proposed by some authors \cite{Lee2019,Tobin2017}. The method presented in \cite{Ghosh2017} suggests partitioning the initial state distribution and train different policies later to be merged in a \highlight{divide-and-conquer} fashion. In highly-stochastic environments, the final policy might be better if the agent does not learn based on the average return but on a \highlight{distribution over returns} \cite{dabney2018implicit,bellemare2017distributional}.

\textbf{Multiagent DRL}\,\,\, In many real-world environments (e.g., robot swarms, autonomous cars), a team of agents must align their behavior while acting in a decentralized way \cite{Rashid2018}; leveraging experience from multiple agents is not always straightforward and other challenges such as partial-observability and non-stationarity might also come into play. An extended review on the subject can be found in \cite{Nguyen2020}. To address this challenge, one approach is to have \highlight{each agent learn independently} \cite{Tampuu2017}, which decentralizes training but might originate stability problems \cite{Foerster2017}. On the opposite side, \cite{foerster2018counterfactual} explores the framework of \highlight{multiple decentralized actors and a single centralized critic}. Inspired by Value Decomposition Networks \cite{sunehag2018value}, the work in \cite{Rashid2018,Son2019} proposes \highlight{hybrid mechanisms} to combine per-agent Q-functions into a single centralized Q-function. \highlight{Multiagent Policy Gradient} algorithms introduce a similar concept focused on continuous spaces \cite{lowe2017multi,li2019robust}, which can be also combined with attention \cite{iqbal2019actor}.

\textbf{Counterfactual reasoning}\,\,\, The ability to reason about actions not taken and ``what-ifs'' is necessary in some real-world systems, especially when constraints or risks are hard to capture. This is a relevant problem in healthcare applications \cite{Prasad2017}. This challenge partly overlaps with offline RL, since extrapolation techniques can be useful in some contexts, especially when there is correlation between state-action pairs inside and outside databases \cite{fujimoto2019off}. While some studies might touch on this concrete challenge, we did not find any work specifically focusing on counterfactuals and real-world DRL. Facebook's platform Horizon \cite{gauci2019horizon}, one of the success stories of real-world DRL, leverages work on Counterfactual Policy Evaluation \cite{wang2017optimal} to evaluate policies without deploying them online.

\textbf{Stability}\,\,\, Once deployed, agents should maintain the desired behavior for indefinite time, even when new experience is collected. This challenge has a link with other considerations: online learning and the problem of catastrophic forgetting, autonomous resets (\cite{Ibarz2021} identifies autonomous resets as one of the specific challenges in the context of robotics), and the general issue of reliability. While this is a challenge directly related to the post-deployment or operation phase, we did not find specific works directly tackling this issue for real-world DRL.

\textbf{Combined challenges}\,\,\, Finally, as pointed out in \cite{Dulac-Arnold2021}, real-world DRL challenges usually do not appear in isolation but combined. The literature specifically addressing multiple challenges simultaneously is scarce. For instance, the work in \cite{Jaderberg2019} focuses on both multiagent settings and partial observability, although they are commonly related. We have not been able to find works tackling numerous challenges at the same time.

\section{Domain-agnostic research's gaps}
\label{sec:gaps}

As seen in Table \ref{tab:summary}, different strategies have been adopted to address specific challenges of real-world DRL; these can be grouped into thirteen types. Based on this literature review, we deem there is a good understanding of each individual challenge, and the provided references demonstrate that novel methods are able to reach new levels of robustness in the test environments. The review has also allowed us to identify the following five gaps that we believe are important to consider when evaluating the progress of real-world DRL.

\textbf{1. Bias towards robotics use cases}\,\,\, Most of the frameworks, use cases, and test environments are focused on control applications, specifically robotics. There are multiple problems in the real world that consist of optimization, design, or recommendation tasks. In some cases their underlying systems are simpler than highly-actuated robots and thus considering them as additional benchmarks could be beneficial moving forward. Extending the focus to these systems might be an opportunity to achieve new successful deployments.

\textbf{2. Not enough research on combined challenges}\,\,\, The majority of the presented studies focus on one challenge at a time and ignore combined challenges analyses. Real-world environments display multiple challenges simultaneously, often with high degrees of interaction. Combined effects are studied in \cite{Dulac-Arnold2021}; their paper proves a simple interaction of a few challenges can substantially hamper the policy performance. They also provide a benchmark task to study this issue in more depth.

\textbf{3. Lack of real-world follow-through}\,\,\, Most of the studies cited in this work make use of the same test environments: MuJoCo \cite{Todorov2012}, DeepMind Control Suite \cite{Tassa2018}, the Arcade Learning Environment \cite{Bellemare2013}, DeepMind Lab \cite{Beattie2016}, Behaviour Suite for RL \cite{Osband2019}, Alchemy \cite{Wang2021}, Meta-World \cite{yu2020meta}, or PyBullet \cite{coumans2016pybullet}. These environments are still simulations; in most cases real-world testing is left as future work. While it is highly difficult to create real-world benchmarks all researchers can use, presenting results on real systems would add value to the community. 

\textbf{4. Little understanding of design tradeoffs}\,\,\, Comparisons of different approaches generally tie loosely to the design perspective and tradeoffs. We believe domain-specific operators looking to introduce DRL into their systems might not find obvious which approaches to try and use when designing their models. Currently, there are little efforts to align all research directions alongside that goal.

\textbf{5. Operation is generally ignored}\,\,\, The vast majority of works ignore operation after deployment, which is an essential piece of information for potential domain-specific operators. In that sense, an open question is: what test scenarios and/or Key Performance Indicators (KPIs) should we use to guarantee operability over time and identify possible performance degradations?

\section{Domain-specific research perspective}

\begin{figure}[t]
\centering
\includegraphics[width=.99\linewidth]{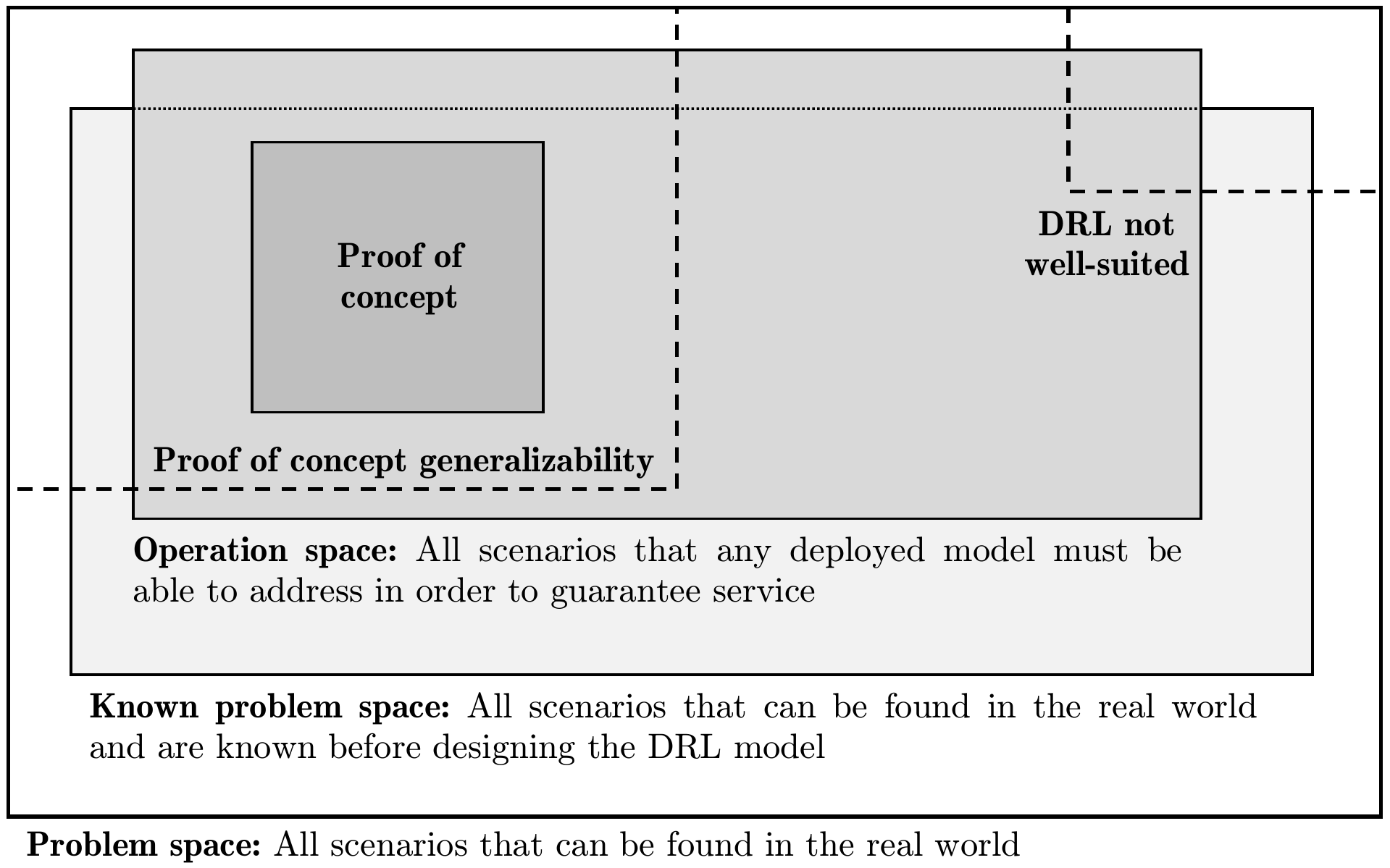}
\caption{Representation of a real-world problem in terms of a problem space, known problem space, and operation space.}
\label{fig:subproblem}
\end{figure}

In the previous sections we outlined the contributions and gaps of RL research when it comes to real-world DRL. In general, though, many of the papers proposing DRL to address concrete problems or applications come from domain-specific communities, mainly following the motivations presented in Section \ref{sec:introduction}. In that sense, the adoption is positive. However, in the majority of cases the path from proofs of concept and prototypes to real deployments and system integrations is still to be traversed. In this section we focus on this issue by, first, reviewing some examples of successful deployments; then, arguing what is missing in other proofs of concept; and finally, outlining possible ways of achieving new deployments that benefit from both RL and domain-specific research.

\subsection{Successful deployments in the real world}

One of the main roadblocks in real-world DRL is training in simulation before deploying. The \textit{sim-to-real gap} is a well-known problem that many applications face. Still, the work in \cite{OpenAI2019} proved a robot could learn manipulation skills and solve a Rubik's Cube only from high-quality simulation training. The agent relies on a simple algorithm, PPO \cite{schulman2017proximal}, to learn the policy. Training in simulation also offered the advantage to easily randomize different environment properties, which helped learning more robustly. \cite{Osinski2020} and \cite{Tai2017} are examples that follow the same framework for autonomous driving and robot mapless navigation, respectively. However, the deployment context in both applications is more complex than solving a Rubik's Cube, therefore the authors only limit themselves to very controlled test conditions in real settings. These differences suggest that the specific real-world problem addressed plays an important role in determining the deployability of a certain DRL model. This motivates our discussion in the following parts of this section.

In some cases, agents might be able to train in the real world and not need to rely on simulations. The work in \cite{Haarnoja2018a} focuses on the ability to learn in a real setting, specifically proposing a method for a quadrupedal robot to learn locomotion skills. By maximizing both return and entropy, the robot acquires a stable gait from scratch in about two hours. Entropy maximization might become essential in real-world training as a means to explore more and better.

In many industries, deployment involves integration within a company's operations. A good example of an industrial deployment is Google Loon's DRL-based station-keeping mechanism \cite{Bellemare2020}. The company replaced its previous controller by a DRL agent that is better at keeping balloons close to base stations. A key feature of this example is that the agent's actions do not alter the environment (i.e., wind currents) and therefore the learning process has less interactions to capture. 

Another relevant industrial deployment is Facebook's Horizon \cite{gauci2019horizon}, which the company uses to decide when to send notifications to users. In this latter case, the possibility to gather massive amounts of data from millions of users has made DRL a successful decision-maker. This mirrors the usefulness of self-play in videogame applications such as AlphaZero \cite{silver2018general} or AlphaStar \cite{vinyals2019grandmaster}.

\subsection{Missing links between proofs of concept and deployments}

We have presented a collection of real-world examples in which DRL is making a difference. Still, the quantity of proofs of concept in the literature substantially outnumbers the quantity of reported deployments. We argue this is due to a combination of different factors inherent to domain-specific research. Throughout this section we use Figure \ref{fig:subproblem} to better understand them.

The premise of the majority of domain-specific papers entails picking a concrete decision-making problem or task of the domain and adapting the DRL framework to address it. In its broadest sense, this problem (e.g., autonomous driving, warehouse management, designing a new drug) can be defined by a \textit{problem space} that captures all possible scenarios that can be encountered in the real world. For example, in the case of autonomous driving, this would correspond to all types of roads, traffic, localizations, etc. Many of these possible scenarios are known beforehand, and therefore constitute the \textit{known problem space}, represented in Figure \ref{fig:subproblem}. We also assume there is a space, whose size varies depending on the application, that encapsulates the scenarios that are unknown to model designers and operators until the model is deployed and/or operated in the real world.

Then, almost all real-world tasks are embedded in business frameworks in which quality of service considerations come into play. This originates an \textit{operation space}, which basically defines what a successful deployment is, especially in industrial contexts. This space is not necessarily completely contained within the known problem space. We argue that in the literature, generally, all proofs of concept of a DRL model addressing a real-world problem in a specific domain are based on scenarios within both the operation and known problem spaces. However, the generalizability of such models is not enough to capture the complete operation space, although it might capture scenarios outside of the operation and the known problem spaces. 

The difference between the operation space and the proofs of concept's generalizability is what is slowing the deployment of DRL models in the real world. Operators might be reluctant to integrate DRL models into the systems until this gap is reduced or completely covered for an individual problem. For instance, in the case of the Rubik's Cube addressed in \cite{OpenAI2019}, the operation space is almost certainly covered by the proposed model, resulting in a successful deployment in real life.

\subsection{Possible considerations to move forward}

The framework presented in Figure \ref{fig:subproblem} is intrinsic to domain-specific research, where methods are outlined within a real problem's context and there is a good understanding of this problem. In contrast, as discussed in Section \ref{sec:gaps}, many of the domain-agnostic studies presented have strong generalizability but lack real-world follow-through, i.e., an operation space as background. When acknowledging both an operation space and the generalizability of proofs of concept, three ideas follow to achieve new deployments: pushing the generalizability boundaries of current proofs of concept, designing additional models that aim to cover remaining areas of the operation space, and considering if certain areas can't be covered by DRL-based methods at all. We weigh in on each of them in the following paragraphs.

A straightforward direction to cover the operation space is to design models that generalize better; this has been a central issue in the RL community. Broadly speaking, the challenges presented in Section \ref{sec:challenges} hamper the generalizability of DRL models in the real-world, and the mitigation strategies discussed in Section \ref{sec:strategies} aim to increase it. However, in domain-specific communities success is in many cases measured by the performance on specific scenarios, ignoring generalizability. Hence, model designers tend to follow different routes and tailor their solutions to very specific scenarios in order to surpass other state-of-the-art methods, usually hand-crafting many elements of the DRL framework in the process. While these dynamics increase engagement in DRL, they generalize poorly and hardly fulfill expectations from the operation perspective. Operators likely care more about generalizability and might want to trade it for performance.

Even though pursuing solutions that generalize better is interesting, it is also fair to assume that a single DRL model might not be sufficient to cover the entire operation space, especially if there are certain quality of service requirements in place. Therefore, another idea to consider is implementing additional proofs of concept so that the overall union covers the operation space. The operator would then need to decide which model to use in which scenario. We believe in some cases this approach is easier to implement than attempting to design a model capable of covering the whole operation space. This is especially easier to ponder from domain-specific research, where an operation space is usually acknowledged. However, current design practices focused on tailoring DRL models to very specific scenarios make this approach hard to scale. The cost of covering the operational space of complex problems by implementing multiple proofs of concept could be unaffordable in some cases.

Finally, we should consider those problems in which no DRL model or combination of models is able to cover the entire operation space. There might be certain operational scenarios that are too complex for a DRL agent (see Figure \ref{fig:subproblem}); we might want to rely on other types of decision-making approaches in those cases. Acknowledging the possible existence of this space could avoid many futile attempts to achieve deployments solely based on DRL, especially if model designers have already explored pushing generalizability boundaries and adding new models. We believe in some cases DRL will not work in the real world only by itself, but in combination with other decision-making technologies.

\section{Conclusion}

In this work we have focused on real-world-oriented Deep Reinforcement Learning (DRL) research from both domain-agnostic and domain-specific perspectives. We have offered our view on why there is a lack of real-world deployments of DRL models despite the numerous efforts from different research communities, and identified different directions to move forward. On one hand, we have provided a comprehensive review of the domain-agnostic challenges of real-world DRL and summarized which are the different approaches being taken to address them. Thanks to this review, we have identified five gaps in domain-agnostic research: a bias towards robotics use cases, not enough research on combined challenges, a lack of real-world follow-through, little understanding of the design tradeoffs, and an omission of operation considerations. On the other hand, we have explained the motivations and success stories of domain-specific research when it comes to DRL. Still, the number of deployments is low. We attribute this to a misalignment between the generalizability of proofs of concept in the literature and operation requirements. Finally, we have discussed possible ways to increase the operability and robustness of domain-specific DRL models and how those can benefit from the research on how to mitigate real-world DRL challenges.



\bibliography{library}
\bibliographystyle{icml2021}


\end{document}